\documentclass[10pt,twocolumn,letterpaper]{article}

\usepackage{cvpr}              

\usepackage{graphicx}
\usepackage{amsmath}
\usepackage{amssymb}
\usepackage{booktabs}
\usepackage{tabu}
\usepackage{float}

\usepackage[pagebackref,breaklinks,colorlinks]{hyperref}

\usepackage[capitalize]{cleveref}
\crefname{section}{Sec.}{Secs.}
\Crefname{section}{Section}{Sections}
\Crefname{table}{Table}{Tables}
\crefname{table}{Tab.}{Tabs.}

\def\cvprPaperID{9033} 
\def\confName{CVPR}
\def\confYear{2023}

\begin{document}

\title{Streaming Video Model}

\author{
Yucheng Zhao$^{1}$\footnotemark[1], Chong Luo$^{2}$, Chuanxin Tang$^{2}$, Dongdong Chen$^{3}$, Noel Codella$^{3}$, Zheng-Jun Zha$^{1}$\footnotemark[2] \\
$^{1}$University of Science and Technology of China \quad $^{2}$Microsoft Research Asia \quad
$^{3}$Microsoft Cloud + AI \\
{\tt\small \{lnc@mail., zhazj\}@ustc.edu.cn \quad \{cluo,chutan,dochen,ncodella\}@microsoft.com}
}

\makeatletter
\g@addto@macro\@maketitle{
  \begin{figure}[H]
  \setlength{\linewidth}{\textwidth}
  \setlength{\hsize}{\textwidth}
  \centering
     \includegraphics[width=0.9\linewidth]{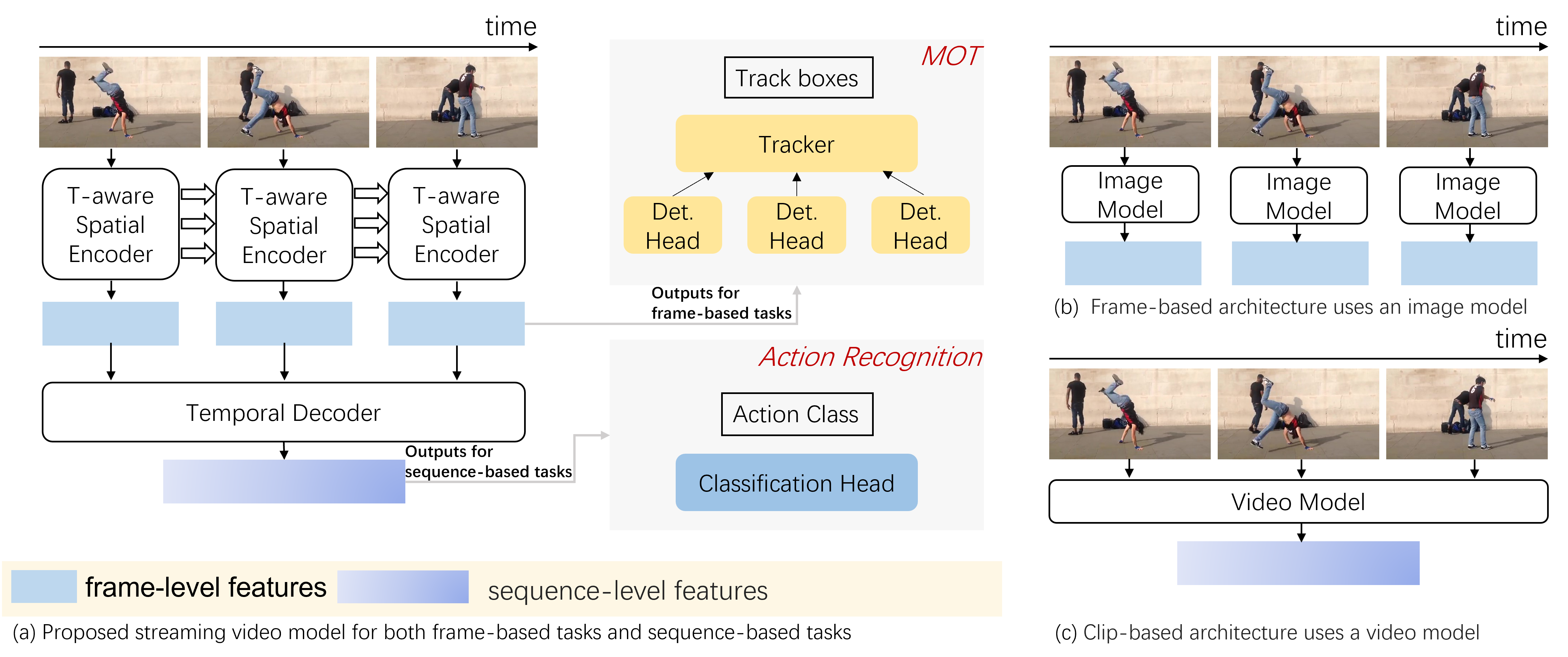}

  \caption{Illustration of the proposed streaming video model with a comparison to conventional frame-based architecture and clip-based architecture. (a) The two-stage streaming video model gracefully serves different types of video tasks through a unified architecture. The output of the temporal-aware (T-aware) spatial encoder serves the frame-based tasks, such as MOT, while the output of the temporal decoder serves the sequence-based tasks, such as action recognition. (b) Frame-based architecture, which uses single image model to independently extract spatial features for each frame, is widely used in the frame-based video tasks. (c) Clip-based architecture, which uses video model to produce the spatiotemporal features for an entire clip, is widely used in the sequence-based video tasks.}
  \label{fig:fig1}
  \end{figure}
}
\makeatother

\maketitle

\renewcommand{\thefootnote}{\fnsymbol{footnote}}
\footnotetext[1]{This work was done during the internship of Yucheng at MSRA.}
\footnotetext[2]{Corresponding author.}
\renewcommand{\thefootnote}{\arabic{footnote}}



\begin{abstract}


Video understanding tasks have traditionally been modeled by two separate architectures, specially tailored for two distinct tasks. Sequence-based video tasks, such as action recognition, use a video backbone to directly extract spatiotemporal features, while frame-based video tasks, such as multiple object tracking (MOT), rely on single fixed-image backbone to extract spatial features. In contrast, we propose to unify video understanding tasks into one novel streaming video architecture, referred to as Streaming Vision Transformer (S-ViT). S-ViT first produces frame-level features with a memory-enabled temporally-aware spatial encoder to serve the frame-based video tasks. Then the frame features are input into a task-related temporal decoder to obtain spatiotemporal features for sequence-based tasks. The efficiency and efficacy of S-ViT is demonstrated by the state-of-the-art accuracy in the sequence-based action recognition task and the competitive advantage over conventional architecture in the frame-based MOT task. We believe that the concept of streaming video model and the implementation of S-ViT are solid steps towards a unified deep learning architecture for video understanding. Code will be available at \url{https://github.com/yuzhms/Streaming-Video-Model}.

\end{abstract}
\section{Introduction}


As a fundamental research topic in computer vision, video understanding mainly deals with two types of tasks. The sequence-based \cite{DBLP:conf/cvpr/CarreiraZ17,DBLP:journals/corr/abs-2209-07526} tasks aim to understand what is happening in a period of time. For example, the action recognition task classifies the object action in a video sequence into a set of predefined categories. The frame-based tasks \cite{DBLP:conf/eccv/ChengS22,DBLP:conf/eccv/ZhangSJYWYLLW22,DBLP:journals/corr/abs-2112-00995}, on the other hand, aim to look for key information in a certain point of time in a video. For example, the multiple object tracking (MOT) task predicts the bounding boxes of objects in each video frame. Although both types of tasks take a video as input, they are handled very differently in computer vision research.    


\begin{figure}[tb]
  \centering
  \includegraphics[width=0.95\linewidth]{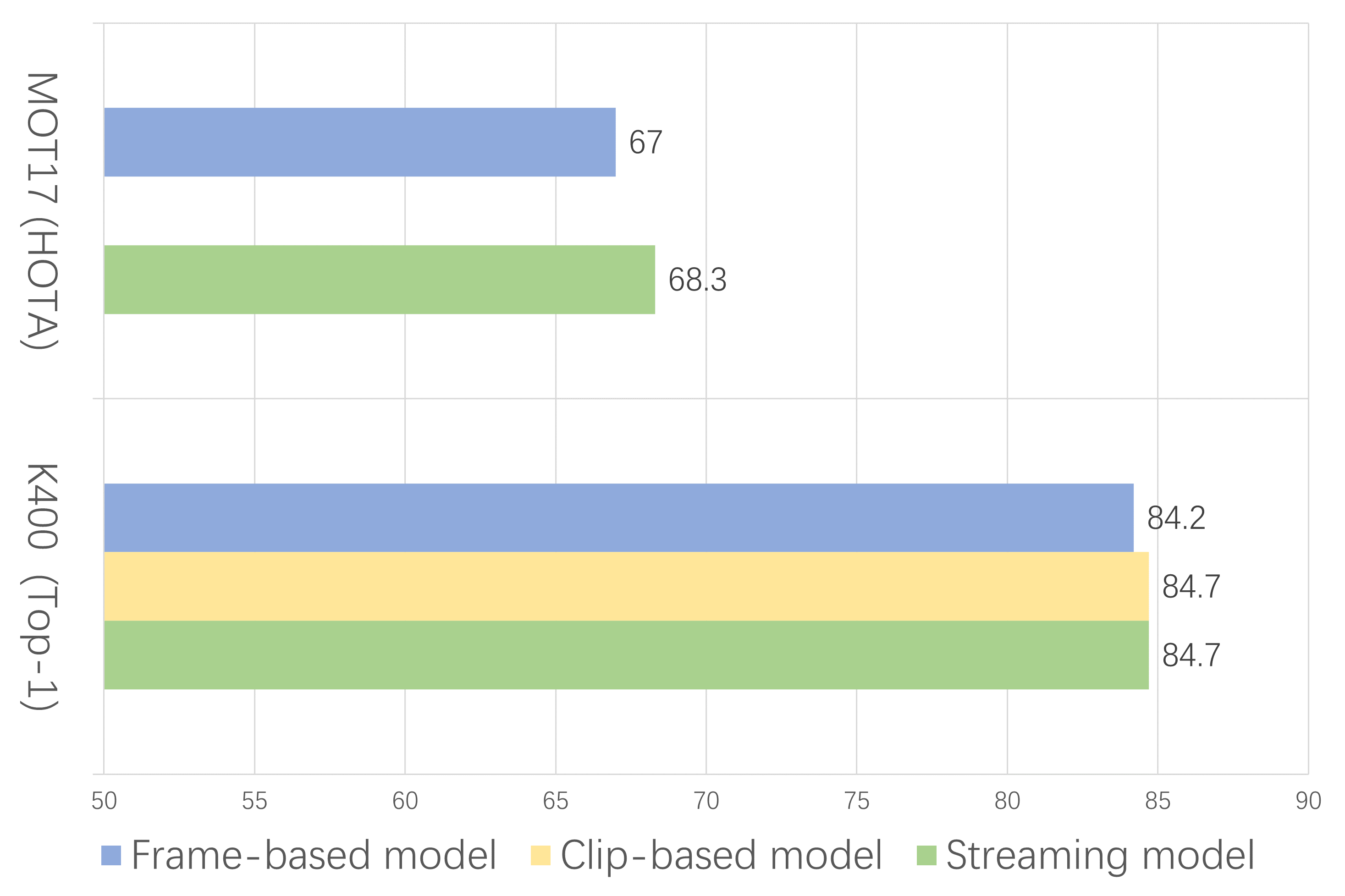}

   \caption{Comparison on video modeling paradigm on both the sequence-based action recognition task and frame-based multiple object tracking task. The proposed streaming model achieves higher performance than the frame-based model on both tasks while has no loss compared to clip-based model on the sequence-based task. The clip-based model can not be directly used in frame-based tasks.}
   \label{fig:fig0}
\end{figure}


The different treatment of these two types of tasks is mainly reflected in the type of backbone network used. The action recognition task is usually handled by a clip-based architecture, where a video model \cite{Arnab_2021_ICCV}, which takes a video clip as input and outputs spatiotemporal features, is used. In the video object segmentation (VOS), video object detection (VOD), and multiple object tracking (MOT) tasks, however, a frame-based architecture \cite{DBLP:conf/cvpr/HeZRS16,DBLP:conf/iclr/DosovitskiyB0WZ21} is often adopted. The frame-based architecture employs image backbone to generate independent spatial features for each frame. In most tracking-by-detection MOT solutions, these features are directly used as the input to the object detector. 

Both types of treatment have their respective drawbacks. On the one hand, the clip-based architecture processes a group of video frames at one time, which puts great pressure on the processor's memory space and processing power. As a result, it is difficult to handle long videos or long actions effectively. In addition, the summarized spatiotemporal features extracted by a video backbone usually lack sufficient spatial resolution to be used for dense prediction tasks. On the other hand, the frame-based architecture does not consider surrounding frames in the process of spatial feature extraction. As a result, the features do not contain any temporal information or an out-of-band mechanism is in need to gather additional temporal information. We believe that a video frame should be treated differently from a single image and that temporal-aware spatial features are more powerful for solving frame-based video understanding tasks.  



In this paper, we propose a unified architecture to handle both types of video tasks. The proposed streaming video model, as shown in Fig.\ref{fig:fig1}, circumvents the drawbacks of the conventional treatment by a two-stage design. Specifically, it is composed of a temporal-aware spatial encoder, which extracts temporal-aware spatial feature for each video frame, and a task-related temporal decoder, which transfers frame-level features to task-specific outputs for sequence-based tasks. 
When compared with frame-based architecture, the temporal-aware spatial encoder in streaming video model leverages additional information from past frames, so that it has potential to obtain more powerful and robust features. When compared with clip-based architecture, our model disentangles the frame-level feature extraction and clip-level feature fusion, so as to alleviate the computation pressure while enabling more flexible use scenarios, such as long-term video inference or online video inference. 


We instantiate such a streaming video model by building the streaming video Transformer (S-ViT) based on the vision Transformer \cite{DBLP:conf/iclr/DosovitskiyB0WZ21}. S-ViT is featured by self-attention within a frame to extract spatial information and cross-attention across frames to make the fused feature temporal-aware. Specifically, for the first frame of a video, S-ViT extracts exactly the same spatial feature as a standard image ViT, but it stores keys and values of every Transformer layer in a memory. For subsequent frames in a video, both intra-frame self-attention and inter-frame cross-attention \cite{DBLP:conf/nips/VaswaniSPUJGKP17} with the stored memory is calculated. S-ViT borrows ideas from triple 2D (T2D) decomposition \cite{zhao2023td} and limits the cross-attention region within patches with the same horizontal or vertical positions. This decomposition reduces the computational cost and allows S-ViT to handle long histories. The output of this stage can directly be used by the frame-based video tasks. For sequence-based tasks, an additional temporal decoder, implemented by a temporal Transformer, is used to gather information from multiple frames. 


We evaluate out S-ViT model on two downstream tasks. The first task is the sequence-based action recognition. We get 84.7\% top-1 accuracy on Kinetics-400 \cite{DBLP:journals/corr/KayCSZHVVGBNSZ17} dataset and 69.3\% top-1 accuracy on Something-Something v2 \cite{DBLP:conf/iccv/GoyalKMMWKHFYMH17} dataset, which is on par with the state-of-the-art, but at a reduced computation expenditure. The second task is MOT, which operates on video frames in a widely adopted tracking-by-detection framework. We show that introducing temporal-aware spatial encoder creates comparative advantage over a frame-based architecture under a fair setting on MOT17 \cite{MOT16} benchmark.


We summarize the contributions as follows. First, we propose a unified architecture, named streaming video model, for both frame-based and sequence-based video understanding tasks. Second, we implement a T2D-based streaming video Transformer and demonstrate how it can be used to serve different types of video tasks. Third, experiments on action recognition and MOT tasks show that our unified model could achieve state-of-the-art results on both types of tasks. We believe that the work presented in this paper is a solid step towards a universal video processing architecture.

\section{Related Works}

\noindent\textbf{Video models and video tasks.} Video understanding is a fundamental research topic in computer vision. There are mainly two kinds of tasks, one of which, named sequence-based tasks \cite{DBLP:conf/cvpr/CarreiraZ17,DBLP:journals/corr/abs-2209-07526}, aims to understand what is happening over a period of time, and the other, named frame-based tasks \cite{DBLP:conf/eccv/ChengS22,DBLP:conf/eccv/ZhangSJYWYLLW22,DBLP:journals/corr/abs-2112-00995}, aims at capture the detail information at a certain point of time. Due to  the fact that the inputs are quite different for these two types of tasks, different families of models are developed independently.

For sequence-based tasks, clip-based models with 3D (width, height, and time) video input are used. 3D convolutional neural networks (CNNs) \cite{DBLP:conf/iccv/TranBFTP15,DBLP:conf/cvpr/CarreiraZ17,DBLP:conf/iccv/QiuYM17,DBLP:conf/cvpr/TranWTRLP18,DBLP:conf/eccv/XieSHTM18,DBLP:conf/iccv/TranWFT19,DBLP:conf/cvpr/Feichtenhofer20} once were popular in the past decade, and video vision Transformer \cite{Arnab_2021_ICCV,DBLP:journals/corr/abs-2106-13230,DBLP:journals/corr/abs-2106-05968,Fan_2021_ICCV,DBLP:conf/icml/BertasiusWT21,zhao2023td,DBLP:journals/corr/abs-2201-04288} are emerging models in recent years. Thanks to the attention mechanism in Transformer \cite{DBLP:conf/nips/VaswaniSPUJGKP17}, video vision Transformers have better capability to model long-range spatiotemporal correlations and thus achieve higher performance than CNN-based methods. For frame-based tasks, frame-based models with 2D (width and height) image input are used. The common models includes ResNet \cite{DBLP:conf/cvpr/HeZRS16}, CSPNet \cite{DBLP:conf/cvpr/WangLWCHY20}, and Swin Transformer \cite{Liu_2021_ICCV}. Such models are adopted in the same way they are for images. And they do not encode any temporal-related information. In this paper, we propose a unified architecture to handle both types of tasks.


\noindent\textbf{Long-term video models and online video models.} As the clip-based models require all frames as input at once, they have difficulty with long videos. A series of works termed long-term video models \cite{DBLP:conf/cvpr/WuLM0XMF22,DBLP:conf/cvpr/WuF0HKG19} are proposed to handle long videos. Building on top of clip-based models, some memory designs are used to extend the temporal coverage. Long-term feature banks \cite{DBLP:conf/cvpr/WuF0HKG19}  augment  3D CNNs with auxiliary supporting information extracted over the entire video. MeMViT \cite{DBLP:conf/cvpr/WuLM0XMF22} augmented multi-scale vision Transformer with cached memories using attention-based designs. There is also a series of methods termed online video models \cite{DBLP:conf/cvpr/KondratyukYLZTB21,DBLP:conf/iccv/LinGH19,DBLP:conf/eccv/ZolfaghariSB18}. The temporal shifted module (TSM) \cite{DBLP:conf/iccv/LinGH19} proposes to shift part of the channels along the temporal dimension to exchange temporal information, resulting in an efficient and online video model. MoViNets \cite{DBLP:conf/cvpr/KondratyukYLZTB21} leverages the neural architecture search (NAS) technique, the causal convolution \cite{DBLP:conf/ssw/OordDZSVGKSK16}, to build an efficient and causal video model for mobile devices.

Our streaming video model does not fall into these two model families as we target unifying frame-based and sequence-based tasks, so it is versatile for any kinds of video inputs. On the contrary, long-term video models and online video models are still clip-based models, where the former aims at extending the temporal context and the latter aims at efficient and causal video model inference. 

\noindent\textbf{Vision Transformer.} Motivated by the success in NLP \cite{DBLP:conf/nips/VaswaniSPUJGKP17}, Vision Transformers (ViTs) \cite{DBLP:conf/iclr/DosovitskiyB0WZ21} have made great progress in computer vision. Different from previous dominant CNN architectures, ViTs treat an image as a set of visual words and model their correlation with the attention operation. ViTs have already led a paradigm shift in various vision tasks, including image recognition \cite{Liu_2021_ICCV}, object detection\cite{DBLP:conf/eccv/CarionMSUKZ20}, semantic segmentation \cite{DBLP:conf/nips/ChengSK21}, action recognition \cite{DBLP:conf/icml/BertasiusWT21}, etc. In this work, we build our streaming video Transformer based on the vanilla vision Transformer \cite{DBLP:conf/iclr/DosovitskiyB0WZ21} and a corresponding video adaptation mechanism triple 2D decomposition \cite{zhao2023td}.

\noindent\textbf{Multiple object tracking.} Tracking by detection \cite{DBLP:conf/icip/BewleyGORU16,DBLP:journals/ijcv/ZhangWWZL21} is one of the dominant paradigms in multiple object tracking (MOT). These method first utilize powerful detectors to obtain detection results in each single frame and then associate defections over time to construct tracking trajectories. The association can be done by using location, motion, and appearance clues or directly solved using transformer architecture as set-prediction \cite{DBLP:conf/eccv/ZengDZWZW22}. We follow a simple yet effective association method called ByteTrack \cite{DBLP:conf/eccv/ZhangSJYWYLLW22} in this paper and use a ViT-based detector to produce detection results. The key feature of the proposed method is the incorporation of a temporal-aware mechanism during the detection feature extraction stage. While some prior works investigate the utilization of temporal information in MOT\cite{DBLP:conf/iccv/Xu0ZH19,DBLP:journals/corr/abs-2104-00194,DBLP:conf/cvpr/ZhouYKK22}, infusing it at the early feature extraction stage is infrequent.

\begin{figure*}[tb]
  \centering
   \includegraphics[width=0.9\linewidth]{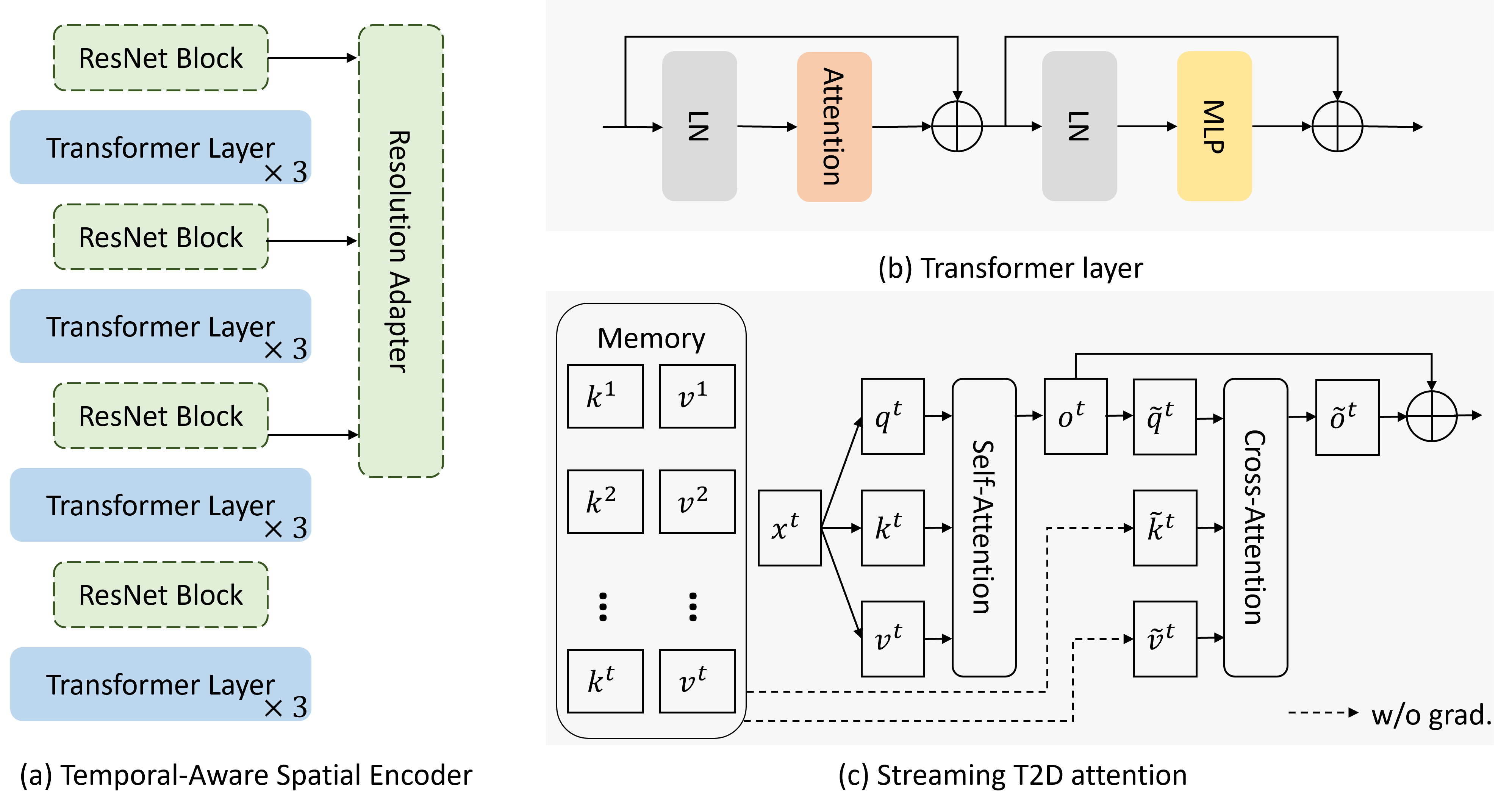}

   \caption{Illustration of streaming video Transformer. (a) The architecture of temporal-aware spatial encoder. (b) The scheme of a Transformer layer. (c) Detailed structure of streaming T2D attention.}
   \label{fig:fig2}
\end{figure*}

\section{Method}

We build a streaming video model, named S-ViT, based on vision Transformer (ViT) \cite{DBLP:conf/iclr/DosovitskiyB0WZ21}. In this section, we will first introduce the background of each S-ViT component. Then, we will describe our architecture and model in details. Finally, we will provide the implementation details. 


\subsection{Background}

Let us first review the vision transformer and its extension to frame-based video tasks and sequence-based video tasks.

\noindent\textbf{Vision Transformer.} Vision Transformer (ViT) is first proposed to process image inputs $X \in \mathbb{R}^{H\times W \times 3}$, where $H$ and $W$ denote the height and width, and 3 is the number of RGB channels. ViT first embeds an image into $N$ non-overlapping patches $X_p\in \mathbb{R}^{N\times C}$, where $C$  is the number of channels. Then, a positional embedding is added to obtain the input $Z^0$ to the first Transformer layer:
\begin{equation}
    Z^0 = X_p + e,
\end{equation}
where $e \in \mathbb{R}^{N \times C}$ is the learnable positional embedding.

The key components in ViT are $L$ Transformer layers which are composed of a self-attention (SA) block, layer normalization (LN) layers, and a multi-layer perception (MLP) block, as shown in Fig.\ref{fig:fig2}-(b). Denote $Z^{l-1}$ and $Z^l$ as the input and output of the $l^{th}$ Transformer layer, the computation implemented by this layer can be written as:
\begin{align}
    Y^l &= \mathrm{MSA}(\mathrm{LN}(Z^{l-1})) + Z^{l-1}, \\
    Z^{l} &= \mathrm{MLP}(\mathrm{LN}(Y^l)) + Y^l. 
\end{align}

\noindent\textbf{ViT for frame-based video tasks.} Most frame-based video tasks, such as VOS, VOD, and MOT, need multi-scale feature maps. ViT is a non-hierarchical architecture that only maintains a single-scale feature map, which makes it difficult to be plugged into existing frameworks.
For example, most detection frameworks utilize the ResNet-style multi-stage architecture that has feature maps of stride 4, 8, 16, and 32, but the plain ViT only has a feature map of stride 16. To solve this resolution misalignment problem, we develop a simple resolution adaptor (RA) to transfer the single-scale feature to multi-scale features, as shown in Fig.\ref{fig:fig2}-(a). The RA is implemented by a set of up-sample and down-sample (de-)convolutions. We found such direct adaptation works well in our video dense prediction tasks. Our implementation is similar to the prior work ViTDet \cite{DBLP:conf/eccv/LiMGH22} that built a simple feature pyramid network (FPN) for image object detection. The difference is that our resolution adaptor does not replace the original sophisticated feature pyramid network (e.g. the PAN \cite{DBLP:conf/cvpr/LiuQQSJ18} in YOLOX \cite{DBLP:journals/corr/abs-2107-08430}) but serves as a plugged-in module on top of the backbone to bridge the resolution mismatch. Besides the multi-scale architecture, plain ViT also has a high computation cost due to the quadratic complexity in self-attention \cite{DBLP:conf/nips/VaswaniSPUJGKP17}. We solve this issue by using windowed self-attention \cite{Liu_2021_ICCV} and convolutional cross-window propagation blocks \cite{DBLP:conf/cvpr/HeZRS16}, which are the same as ViTDet.

\noindent\textbf{ViT for sequence-based video tasks.} Classical clip-based video models need to model the spatiotemporal feature jointly. Although our model does not follow the clip-based paradigm, the spatiotemporal feature learning mechanics in existing works are profitable in the streaming model's design. From this perspective, we build our streaming video model from a SOTA clip-based video model named T2D-ViT \cite{zhao2023td}. The T2D-ViT extends ViT from an image model to a clip-based video model by introducing temporal attention. Concretely, given an input video tensor $Z \in \mathbb{R}^{N_h\times N_w \times N_t \times C}$, besides calculating the XY attention inside each frame, T2D-ViT also calculates the XT temporal attention within the same $y\in \{1,2,..., N_h\}$ index and the TY temporal attention within the same $x\in \{1,2,..., N_w\}$ index.

The central idea of T2D-ViT is the decomposition of static appearance and dynamic motion. Therefore, it shares the same spirit with our streaming video model in that the spatial modeling in the current frame and the temporal modeling among nearby frames are disentangled. Due to the efficiency and effectiveness of T2D-ViT, we adopt a similar XT and TY temporal attention in our S-ViT model, which will be introduced in the next section.

\subsection{Streaming Video Model}

Fig.\ref{fig:fig1}-(a) gives an overview of the proposed streaming video Transformer. Given an input sequence, at each timestamp, the temporal-aware spatial encoder module first encodes the spatial information within the current frame; then it fuses information from previous timestamps. The output of this module is frame-level features, which can be utilized for frame-based tasks like multiple object tracking. On top of the temporal-aware spatial encoder, an optional temporal decoder is appended to generate video-level features. Such video-level features are used for sequence-based tasks like action recognition.

The core design in our streaming video Transformer is the temporal-aware spatial encoder with streaming T2D attentions. The architecture of temporal-aware spatial encoder is shown in Fig.\ref{fig:fig2}-(a), which is composed of multiple Transformer layers and optional ResNet Blocks and the resolution adaptor. The ResNet Block and the resolution adaptor are used for frame-based video tasks which needs multi-scale feature maps. The Transformer layer is composed of Attention layer and MLP block with skip connection and layer normalization, as shown in Tab.\ref{fig:fig2}-(b). We use the streaming T2D attention, which introducing temporal-aware spatial features by leveraging memorized histories.


Fig.\ref{fig:fig2}-(c) illustrates the implementation of streaming T2D attention. First, we compute the spatial self-attention from the input $x_t$:
\begin{align}
    q_t&=x_t W_q; k_t=x_t W_k; v_t=x_t W_v, \\
    o_t&=\mathrm{Attention}(q_t, k_t, v_t),
\end{align}
where $W_q$, $W_k$, and $W_v$ are projection matrices for queries, keys and values, respectively. Then, we maintain a memory pool to store the historical information. During each frame's forward process, we put the keys and values in the self-attention into the memory pool. Concretely, in the forward pass of the first frame, the memory pool only contains the keys and values of the first frame itself. And in the forward pass of the t-th frame, the memory pool contains all keys and values from the past timestamps. Formally speaking, the memory used for frame t is 
\begin{align}
\Tilde{k}^{t} &= [\mathrm{sg}(k^1), \mathrm{sg}(k^2), ..., \mathrm{sg}(k^{t-1}), \mathrm{sg}(k^t)], \\
\Tilde{v}^{t} &= [\mathrm{sg}(v^1), \mathrm{sg}(v^2), ..., \mathrm{sg}(v^{t-1}), \mathrm{sg}(v^t)].
\end{align}
Here the $\mathrm{sg}$ stands for \texttt{stop gradient}. We generate another temporal query $\Tilde{q}^{t}$ from the output of spatial self-attention $o_t$ by a separate transformation matrix $\Tilde{W}_q$ and then compute the cross attention on $\Tilde{q}^{t}$, $\Tilde{k}^{t}$, and $\Tilde{v}^{t}$:
\begin{equation}
    \Tilde{o}^t = \mathrm{Attention}(\Tilde{q}_t, \Tilde{k}_t, \Tilde{v}_t).
\end{equation}
Notice that the cross-attention here is calculated within the XT and TY data planes to improve the efficiency and effectiveness. Using the TY attention as an example. Given inputs $\Tilde{q}_t \in \mathbb{R}^{ N_w\times N_t \times C}$ and $\Tilde{k}_t,\Tilde{v}_t \in \mathbb{R}^{ T\times N_w\times N_t \times C}$, we split them along the horizontal axes to get $\{\Tilde{q}_t^1, \Tilde{q}_t^2, ..., \Tilde{q}_t^{N_w}\}$,  $\{\Tilde{k}_t^1, \Tilde{k}_t^2, ..., \Tilde{k}_t^{N_w}\}$, and  $\{\Tilde{v}_t^1, \Tilde{v}_t^2, ..., \Tilde{v}_t^{N_w}\}$. The attention is calculated among queries, keys, and values with the same horizontal axis. Similarly, XT attention is calculated among queries, keys, and values with the same vertical axis. The outputs of XT and TY attention are fused into $o_t$ with learnable per-channel weights initialized to $1e-4$. The introduction of T2D attention decrease the computational complexity of cross attention part from $O(N_w^2N_h^2T)$ to $O(N_w^2N_hT + N_w N_h^2T)$, which makes our temporal attention module light-weight and therefore applicable for long histories.

\subsection{Implementation Details}

We implement our S-ViT based on the ViT-B \cite{DBLP:conf/iclr/DosovitskiyB0WZ21} model, which has 12 layers of Transformer. To support multi-scale features, we manually split the network into 4 stages, with 3 layers for each stage. In the frame-based video tasks, we use windowed attention with the window size of $14\times 14$ to reduce the heavy computation cost from the global self-attentions. Four ResNet blocks are appended at the end of each stage, respectively, for cross-window feature propagation. The CLIP \cite{DBLP:conf/icml/RadfordKHRGASAM21} pre-trained weights are used as the initialization. For parameters that did not exist in the ViT-B model, we randomly initialized them. 

For the action recognition task, we use four temporal transformer layers as the temporal decoder. A text-generated classifier \cite{DBLP:journals/corr/abs-2207-01297} is applied for the Kinetics-400 \cite{DBLP:journals/corr/KayCSZHVVGBNSZ17} dataset and a learnable linear classifier for the something-something v2 \cite{DBLP:conf/iccv/GoyalKMMWKHFYMH17} dataset following T2D-ViT. On the multiple object tracking task, we use the YOLOX-style \cite{DBLP:journals/corr/abs-2107-08430} detection head and the ByteTrack \cite{DBLP:conf/eccv/ZhangSJYWYLLW22} tracker.
\input\section{Experiments}

\subsection{Experimental Setup}
We evaluate our method on two video tasks, namely the video action recognition and the multiple object tracking. For video action recognition, we conduct experiments on two widely used benchmark, i,e., Kinetics-400 \cite{DBLP:journals/corr/KayCSZHVVGBNSZ17} and Something-Something v2 \cite{DBLP:conf/iccv/GoyalKMMWKHFYMH17}. For multiple object tracking, we use MOT17 \cite{MOT16} dataset for evaluation with additional data sources MOTSynth \cite{DBLP:conf/iccv/FabbriBMCGOCLC21} and CrowdHuman \cite{DBLP:journals/corr/abs-1805-00123}.

\textbf{Kinetics-400 (K400) \cite{DBLP:journals/corr/KayCSZHVVGBNSZ17}} is a large-scale video action recognition  dataset collected from YouTube. It contains 234584 training videos and 19760 validation videos. The video in K400 is trimmed to around 10 seconds. We use the sparse sampling \cite{DBLP:journals/corr/abs-2208-02816} and randomly resized cropping to sample 16 frames with $224\times 224$ resolution to form a video clip. We use the same data augmentation and regularization as in X-CLIP \cite{DBLP:journals/corr/abs-2208-02816}, including random horizontal flip, color jitter, random grayscale, label smoothing, Mixup \cite{DBLP:conf/iclr/ZhangCDL18}, and CutMix \cite{DBLP:conf/iccv/YunHCOYC19}. In the inference phase, we adopt the multi-view testing with four temporal clips and three spatial crops. The top-1 and top-5 classification accuracy on the validation set are reported as evaluation metrics. 

\textbf{Something-Something V2 (SSv2) \cite{DBLP:conf/iccv/GoyalKMMWKHFYMH17}} is another large-scale action recognition dataset which focus more on temporal modeling. The labels are like "Pulling something from left to right", so it is crucial to learn motion information. The training set contains 168.9K training videos and the validation set contains 24.7K validation videos. We use segment-based sampling from \cite{DBLP:conf/iccv/LinGH19} to sample 32 frames with $224\times 224$ resolution. The augmentation and regularization in SSv2 include random augmentation \cite{DBLP:conf/nips/CubukZS020}, repeated augmentation \cite{DBLP:conf/cvpr/HofferBHGHS20}, random erasing \cite{DBLP:conf/aaai/Zhong0KL020}, Mixup \cite{DBLP:conf/iclr/ZhangCDL18}, and CutMix \cite{DBLP:conf/iccv/YunHCOYC19}, which follow the practice in MViT \cite{Fan_2021_ICCV}.

\begin{table*}[tbh!]
    \centering
    \caption{Comparison of the frame-based video model and streaming video model on MOT17 half-validation set.}
    \begin{tabular}{@{}l|c|c|c|c|c|c@{}}
        \toprule
        Method  & MOTA $\uparrow$& IDF1 $\uparrow$ & HOTA $\uparrow$ & FP $\downarrow$ & FN $\downarrow$ & IDs $\downarrow$\\
        \hline
        frame-based  &79.0 & 78.4 & 67.0 & 10248 & 23058 & 564    \\
        streaming   & 79.6 & 80.9 & 68.3 & 9507  & 22956 & 453\\
        \bottomrule
    \end{tabular}
    \label{table1}
\end{table*}

\begin{figure}[tb]
  \centering
  \includegraphics[width=0.95\linewidth]{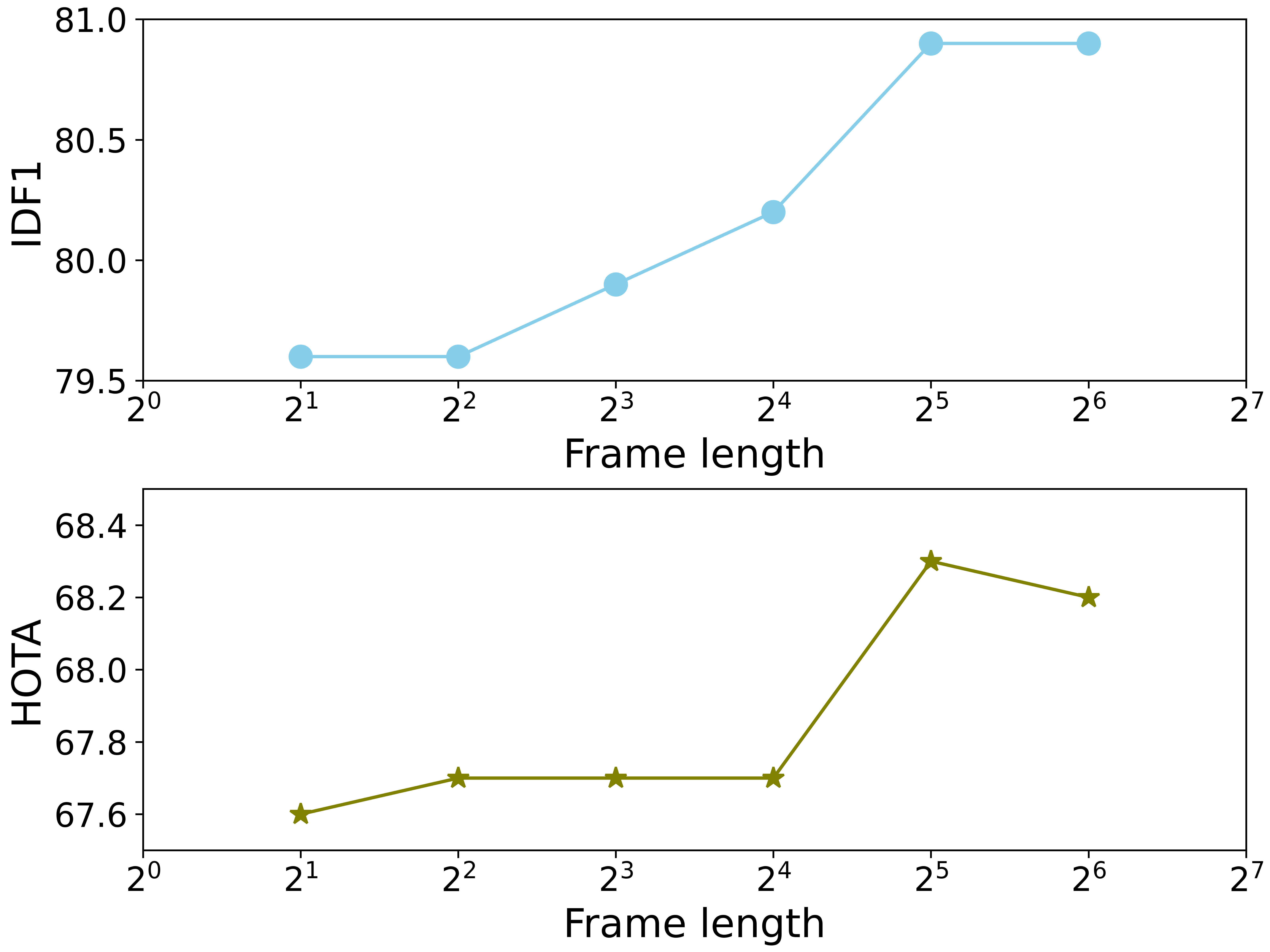}

   \caption{Comparison of the performance of S-ViT with different test-time memory length.}
   \label{fig:fig4}
\end{figure}

\begin{table}[tb]
    \centering
    \caption{Comparison of training datasets for streaming video model training on MOT17 half-validation set. MOT17 is a video dataset. MOTS is short for MOTSynth, which is a synthetic video dataset. CH is short for CrowdHuman, which is an image dataset.}
    \resizebox{\linewidth}{!}{
    \begin{tabular}{@{}l|c|c|c|c|c|c@{}}
        \toprule
        Dataset  & MOTA $\uparrow$& IDF1 $\uparrow$ & HOTA $\uparrow$  & FP $\downarrow$ & FN $\downarrow$ & IDs $\downarrow$\\
        \hline
        
        MOT17                                   & 69.9 & 73.6 & 61.6 & 18837 & 29016 & 750 \\
        +CH                  & 78.0 & 78.0 & 65.7 & 9966  & 25065 & 549 \\
        +MOTS                    & 77.4 & 78.1 & 66.2 & 11742 & 24279 & 555 \\
        +MOTS +CH  & 79.6 & 80.9 & 68.3 & 9507  & 22956 & 453 \\

        \bottomrule
    \end{tabular}}
    \label{table2}
\end{table}

\textbf{MOT17 \cite{MOT16}} is a multiple object tracking dataset that contains 7 training sequences and 7 test sequences. The total frame number is only 11k, so it is not enough to train our S-ViT model. We use the CrowdHuman \cite{DBLP:journals/corr/abs-1805-00123} dataset and the MOTSynth \cite{DBLP:conf/iccv/FabbriBMCGOCLC21} dataset to expand the training data. CrowdHuman contains 19.4k images in crowd human scenarios, and MOTSynth contains 764 synthetic video sequences with 1.3m frames generated from Grand Theft Auto V. We conduct our experiments with combinations of different data sources and discuss the influence in Sec.\ref{sec:4.2}. The data augmentation and regularization include Mosaic \cite{DBLP:journals/corr/abs-2004-10934} and Mixup \cite{DBLP:conf/iclr/ZhangCDL18}, which follow the practice in ByteTrack. The input image size is $1440\times 800$ with the shortest side ranging from 576 to 1024 during multi-scale training. We use the CLEAR \cite{DBLP:journals/ejivp/BernardinS08} metrics for evaluation, including multiple object tracking accuracy (MOTA), high order tracking accuracy (HOTA) \cite{DBLP:journals/ijcv/LuitenODTGLL21}, and IDF1, to evaluate different aspects of tracking and detection performance. We also report the raw statics such as FP, FN, and IDs. As there are no labels for the testing set of MOT17, we split the training set by using the first half of each video for training and the last half for validation in our ablation studies, following \cite{DBLP:conf/eccv/ZhouKK20}. We report test results when compared with other methods.

\textbf{Training configurations.} We train our S-ViT model using the AdamW \cite{DBLP:conf/iclr/LoshchilovH19} optimizer. The training epoch for action recognition on K400 and SSv2 is set to 30 with 5 epochs of warmup. A cosine learning rate schedule with the maximum learning rates of 1e-5 and 5e-5 are used for K400 and SSv2 respectively. The training epoch for multiple object tracking is set to 10 with 1 epoch of warmup. The learning rate is set to 2.5e-4 with a cosine annealing schedule. More details can be found in the supplementary.

\subsection{Results on Multiple Object Tracking}
\label{sec:4.2}

The most important advantage of our S-ViT model for frame-based video tasks is its ability to extract temporal-aware spatial features. We design controlled experiments on the MOT17 dataset to demonstrate the effectiveness of the streaming video model and also ablate the influence of some newly introduced factors.

\noindent\textbf{Effectiveness of streaming video model.} Tab.\ref{table1} shows the comparison between our streaming video model and the frame-based video model. Our streaming video model outperforms the frame-based video model by 0.6 MOTA, 2.5 IDF1, and 1.3 HOTA, which clearly demonstrates the effectiveness of temporal-aware spatial features.

\noindent\textbf{Influence of test-time memory length.} One flexibility of our streaming video Transformer is that we can use arbitrary memory length in the test phase without model re-training. Intuitively, using longer history helps our model to extract robuster features. As shown in Fig.\ref{fig:fig4}, longer test-time memory length indeed improves the tracking performance. Specifically, the 32-frame model gets 1.3 higher IDF1 and 0.7 higher HOTA than the 2-frame model.

\noindent\textbf{Ablation study on training datasets.} The paradigm switch from a frame-based video model to a streaming video model also involves a transition of training datasets. In multiple object tracking, it is common practice to involve additional data sources, as the MOT17 only has seven training sequences. An image pedestrian detection dataset called CrowdHuman is used in many prior works. However, as the CrowdHuman dataset only contains still images, it cannot provide useful temporal information, which is needed by our streaming video model. We thus introduce another synthetic video dataset, MOTSynth, to train our streaming video model. The drawback of using MOTSynth is that it has a domain gap with real images because it is generated from a video game. We evaluate the different combinations of these data sources and present the results on Tab.\ref{table2}. The first row shows the results of using MOT17 alone. The HOTA of this model is only 61.6\% and we observe severe over-fitting during training. The second row and the third row show the results of adding CrowdHuman and MOTSynth, respectively. In order to use CrowHuman in our streaming video model, we duplicate frames to form a video. It is clear that both additional data sources help our model achieving higher performance on MOT17. Finally, in the last row, we use all three data sources and achieve the highest performance of 68.3 HOTA, showing the importance of using both the video data sources and the real-world data sources.

\subsection{Results on Action Recognition}

We present the action recognition results of S-ViT on Tab.\ref{table3} with a comparison of the frame-based model and the clip-based model. The frame-based model here is implemented with a spatial encoder and a temporal decoder, and the streaming model further upgrades the spatial-encoder's temporal awareness. The only difference between them is whether to use temporal awareness in spatial encoder. Our streaming video model achieves a 0.5\% top-1 accuracy gain and a 1.0\% top-1 accuracy gain over the frame-based model on K400 and SSv2, respectively, thanks to the temporal-aware spatial encoder. It is also surprising to see that our streaming video model achieves similar top-1 and top-5 accuracy on K400 when compared with the clip-based model but reduces the GFLOPs by 14\%. The streaming video model only uses the history information to compute the cross-attention, while the clip-based video model uses both the history and the future. The same performance of these two models indicates that future information may not be necessary for sequence-based video task, and we have the opportunity to build a causal video model without sacrificing the performance on some kind of dataset. On SSv2, we observe a notable performance loss that may be related to the fine-grained category definition in SSv2. For example, knowing future information may help to distinguish the class "opening something" from the class "pretending to open something without actually opening it."

\begin{table}[b]
    \centering
    \caption{Comparison of the frame-based video model, clip-based video model, and streaming video model on K400 and SSv2.}
    \begin{tabular}{@{}l|c|c|c|c|c@{}}
        \toprule
        Method  & GFLOPs& \multicolumn{2}{c}{K400} &  \multicolumn{2}{c}{SSv2} \\
        && Top-1 & Top-5 & Top-1 & Top-5 \\
        \hline
        frame-based & 282 & 84.2 & 96.7 & 68.3 & 91.6 \\
        clip-based  & 397 & 84.7 & 96.7 & 70.5 & 92.6\\
        streaming   & 340 & 84.7 & 96.8 & 69.3 & 92.1\\
        \bottomrule
    \end{tabular}
    \label{table3}
\end{table}

\begin{table}[tb]
  \centering
  \caption{Comparison to the state-of-the-art on Kinetics-400. \#Frames denotes the total number of frames used during inference which is \#frames per clip $\times$ \# spatial crop $\times$ \# temporal clip.}
\resizebox{\linewidth}{!}{
    \begin{tabu}{@{}lrrcc@{}}
    \toprule
    Method                    & \#Frames  & GFLOPs& Top-1 & Top-5 \\
    \hline
    \multicolumn{5}{c}{\textit{Methods with CNN}}\\

    R(2+1)D \cite{DBLP:conf/cvpr/TranWTRLP18}                    & 16x1x10   & 75    & 72.0  & 90.0 \\
    SlowFast + NL \cite{DBLP:conf/iccv/Feichtenhofer0M19}                   & 16x3x10   & 234   & 79.8  & 93.9 \\
    X3D-XXL \cite{DBLP:conf/cvpr/Feichtenhofer20}                        & 16x3x10   & 144   & 80.4  & 94.6 \\
    \hline
    \multicolumn{5}{c}{\textit{Methods with Transformer}}\\
    TokenLearner \cite{DBLP:conf/nips/RyooPADA21}             & 64x3x4    & 4076  & 85.4  & 96.3 \\
    ViViT-L FE \cite{Arnab_2021_ICCV}              & 32x3x1    & 3980  & 83.5  & 94.3\\
    MViTv2-L $(312\uparrow)$ \cite{DBLP:journals/corr/abs-2112-01526}    & 40x3x5    & 2828  & 86.1  & 97.0\\
    TimeSformer-L \cite{DBLP:conf/icml/BertasiusWT21}              & 96x3x1    & 2380  & 80.7  & 94.7\\
    Video-Swin-L $(384\uparrow)$ \cite{DBLP:journals/corr/abs-2106-13230}& 32x5x10   & 2107  & 84.9  & 96.7\\
    MTV-L \cite{DBLP:journals/corr/abs-2201-04288}                    & 32x3x4    & 1504  & 84.3  & 96.3\\
    MTV-B \cite{DBLP:journals/corr/abs-2201-04288}  &32x3x4 & 399 & 81.8 & 95.0 \\
    Uniformer-B \cite{DBLP:conf/iclr/Li00S00022}                & 32x3x4    & 259   & 83.0  & 95.4\\
    MViTv2-B \cite{DBLP:journals/corr/abs-2112-01526}    & 32x1x5    & 225   & 82.9  & 95.7 \\
    Video-Swin-S \cite{DBLP:journals/corr/abs-2106-13230}    & 32x3x4    & 166 & 80.6 & 94.5 \\
    \hline
    \multicolumn{5}{c}{\textit{Methods with CLIP-B pre-trained ViT}}\\
    
    ActionCLIP-B/16 \cite{DBLP:journals/corr/abs-2109-08472}              & 32x3x10   & 563   & 83.8  & 96.2\\
    EVL ViT-B/16 \cite{DBLP:journals/corr/abs-2208-03550}                  & 32x3x1    & 592   & 84.2  & - \\
    X-CLIP-B/16 \cite{DBLP:journals/corr/abs-2208-02816}                  & 16x3x4    & 287   & 84.7  & 96.8\\
    ViT-B w/ ST-Adapter \cite{DBLP:journals/corr/abs-2206-13559}          & 32x3x1    & 607   & 82.7  & 96.2 \\
    Text4Vis-B/16 \cite{DBLP:journals/corr/abs-2207-01297}                & 16x3x4    & -     & 83.6  & 96.4 \\
    T2D-B \cite{zhao2023td}             & 16x3x4    & 395   &84.7 & 96.7\\
    \hline
    \multicolumn{5}{c}{\textit{\textcolor{blue}{Streaming Video Model}}}\\
    S-ViT (\textbf{Ours})               & 16x3x4    & 340 &84.7 & 96.8\\

    \hline
  \end{tabu}}
  \label{table4}
\end{table}

\subsection{Benchmark Evaluation}

In this section, we compare the performance of S-ViT with state-of-the-art methods on both the action recognition task and the multiple object tracking task. Results of action recognition on K400 and SSv2 are shown in Tab.\ref{table4} and Tab.\ref{table5}, respectively. Results of multiple object tracking on the MOT17 test set are shown in Tab.\ref{table6}.

\textbf{Kinetics-400.} In Tab.\ref{table4}, we report the comparison of our streaming video model and previous clip-based models on K400. Among all the compared models, our S-ViT achieves competitive performance with relatively low GFLOPs. Specifically, we get a 2.9\% top-1 accuracy gain over MTV-B \cite{DBLP:journals/corr/abs-2201-04288} and a 0.8\% top-1 accuracy gain over EVL ViT-B/16 \cite{DBLP:journals/corr/abs-2208-03550} with lower GFLOPs. Even compared with the state-of-the-art models X-CLIP-B/16 \cite{DBLP:journals/corr/abs-2208-02816} and T2D-B \cite{zhao2023td}, our S-ViT still gets competitive performance. It is worth noting that our model is a streaming video model that extracts features frame by frame and does not use future information in the temporal-aware spatial encoder. So it is quite a success that we do not lag behind clip-based video models, showing the opportunity of using streaming video models on sequence-based tasks.

\textbf{Something-Something V2.} Tab.\ref{table5} presents results of S-ViT compared to SOTA methods on SSv2. Consistent with our findings on K400, our streaming video model showcases considerable proficiency on this motion-focused dataset, demonstrating its potential to operate as a general video action recognition model for diverse datasets.

\textbf{MOT17.} We report multiple object tracking results on MOT17, as shown in Tab.\ref{table6}. Among all the compared methods, our S-ViT attains top performance and only underperforms ByteTrack, which utilizes the strong YOLOX detector with COCO \cite{DBLP:conf/eccv/LinMBHPRDZ14} pre-training. S-ViT uses a pure ViT backbone and does not use any detection pre-training. Further tuning of the ViT-based detection architecture may improve the performance of our method, but it is beyond the scope of streaming video model in this paper. Nevertheless, our S-ViT achieves the highest performance among all Transformer-based methods, outperforming TransMOT \cite{DBLP:journals/corr/abs-2104-00194} by 1.4 MOTA, 0.8 IDF1, and 0.3 HOTA.

\begin{table}[tb]
  \centering
  \caption{Comparison to the state-of-the-art on SSv2.}
  \resizebox{\linewidth}{!}{
    \begin{tabu}{@{}lcccc@{}}
    \toprule
    Method                   & \#Frames  & GFLOPs& Top-1 & Top-5 \\
    \hline
    \multicolumn{5}{c}{\textit{Methods with CNN}}\\
    TSM \cite{DBLP:conf/iccv/LinGH19}                         & 16x1x1    & 66    & 63.3  & 88.5 \\
    MSNet \cite{DBLP:conf/eccv/KwonKKC20}                      & 16x1x1    & 67    & 64.7  & 89.4 \\
    SELFYNet \cite{Kwon_2021_ICCV}                  & 16x1x1    & 67    & 65.7  & 89.8 \\
    TDN \cite{DBLP:conf/cvpr/0002TJW21}                      & 16x1x1    & 132   & 66.9  & 90.9 \\
    \hline
    \multicolumn{5}{c}{\textit{Methods with hierarchical Transformer}}\\

    Video-Swin-B \cite{DBLP:journals/corr/abs-2106-13230}              & 32x3x1    & 321   & 69.6  & 92.7 \\
    UniFormer-B \cite{DBLP:conf/iclr/Li00S00022}               & 32x3x1    & 259   & 71.2  & 92.8 \\
    MViT-B-24 \cite{Fan_2021_ICCV}                  & 32x3x1    & 236   & 68.7  & 91.5 \\
    MViTv2-S \cite{DBLP:journals/corr/abs-2112-01526} & 32x3x1 & 65 & 68.2 & 91.4 \\
    MViTv2-B \cite{DBLP:journals/corr/abs-2112-01526}                    & 32x3x1    & 225   & 72.1  & 93.4 \\
    \hline
    \multicolumn{5}{c}{\textit{Methods with cylindrical Transformer}}\\
    TimeSformer-HR \cite{DBLP:conf/icml/BertasiusWT21}            & 16x3x1    & 1703  & 62.5  & - \\
    ViViT-L \cite{Arnab_2021_ICCV}                     & 16x3x4    & 903   & 65.4  & 89.8 \\
    MTV-B (320p) \cite{DBLP:journals/corr/abs-2201-04288}                & 16x3x4    & 930   & 68.5  & 90.4 \\
    Mformer-L \cite{DBLP:conf/nips/PatrickCAMMFVH21}                  & 32x3x1    & 1185  & 68.1  & 91.2 \\
    EVL ViT-B/16 \cite{DBLP:journals/corr/abs-2208-03550}               & 32x3x1    & 682   & 62.4  & - \\
    ViT-B w/ ST-Adapter \cite{DBLP:journals/corr/abs-2206-13559}        & 32x3x1    & 652   & 69.5  & 92.6 \\
    T2D-B  \cite{zhao2023td}           & 32x3x2    & 397   & 70.5  & 92.6 \\
    \hline
    \multicolumn{5}{c}{\textit{\textcolor{blue}{Streaming Video Model}}}\\
    S-ViT \textbf{(Ours)}            & 32x3x2    & 340   & 69.3  &  92.1 \\
    \bottomrule
  \end{tabu}}
  \label{table5}
\end{table}

\begin{table}[tb]
  \centering
  \caption{Comparison to the state-of-the-art on the MOT17 test set.}
  \resizebox{\linewidth}{!}{
    \begin{tabu}{@{}lcccccc@{}}
    \toprule
    Method & MOTA $\uparrow$& IDF1 $\uparrow$ & HOTA $\uparrow$ & FP $\downarrow$ & FN $\downarrow$ & IDs $\downarrow$\\
    \hline
    \multicolumn{7}{c}{\textit{Methods with CNN}}\\

    CenterTrack \cite{DBLP:conf/eccv/ZhouKK20}&67.8 & 64.7 & 52.2  & 18,498    & 160,332   & 3,039\\
    QDTrack \cite{DBLP:conf/cvpr/PangQLCLDY21}    &68.7 & 66.3 & 53.9  & 26,589    & 146,643   & 3,378\\
    TraDeS \cite{DBLP:conf/cvpr/WuCS00Y21}     &69.1 & 63.9 & 52.7  & 20,892    & 150,060   & 3,555\\
    FairMOT \cite{DBLP:journals/ijcv/ZhangWWZL21}    &73.7 & 72.3 & 59.3  & 27,507    & 117,477   & 3,303\\
    CorrTracker \cite{DBLP:conf/cvpr/WangZPX21}&76.5 & 73.6 & 60.7  & 29,808    & 99,510    & 3,369 \\
    Unicorn \cite{DBLP:conf/eccv/YanJSWYLL22}    &77.2 & 75.5 & 61.7  & 50,087    & 73,349    & 5,379\\
    ByteTrack \cite{DBLP:conf/eccv/ZhangSJYWYLLW22}  &80.3 & 77.3 & 63.1  & 25,491    & 83,721    & 2,196\\
    \hline
    \multicolumn{7}{c}{\textit{Methods with Transformer}}\\

    MeMOT \cite{DBLP:conf/cvpr/CaiX0XXTS22}      &72.5 & 69.0 & 56.9  & 37,221    & 115,248   & 2,724\\
    TransCenter \cite{DBLP:journals/corr/abs-2103-15145}&73.2 & 62.2 & 54.5  & 23,112    & 123,738   & 4,614\\
    MOTR \cite{DBLP:conf/eccv/ZengDZWZW22}       &73.4 & 68.6 & 57.8  & -         & -         & 2,439\\ 
    Trackformer \cite{DBLP:conf/cvpr/MeinhardtKLF22} &74.1 & 68.0 & -     & 34,602    & 108,777   & 2,829\\
    TransTrack \cite{DBLP:journals/corr/abs-2012-15460}  &75.2 & 63.5 & 54.1  & 50,157    & 86,442    & 3,603\\ 
    GTR \cite{DBLP:conf/cvpr/ZhouYKK22}        &75.3 & 71.5 & 59.1  & 26,793    & 109,854   & 2,859 \\
    TransMOT \cite{DBLP:journals/corr/abs-2104-00194}    &76.7 & 75.1 & 61.7  & 36,231    & 93,150    & 2,346\\
    S-ViT (\textbf{Ours}) & 78.1 & 75.9 & 62.0 & 39,063 & 82,704 & 1,983\\

    \bottomrule
  \end{tabu}}
  \label{table6}
\end{table}
\section{Conclusion}

In this work, we propose the idea of streaming video models that aim to unify the treatment of both frame-based and sequence-based video understanding tasks, which in the past were handled by separate models. We present an implementation named streaming video Transformer and conduct comprehensive experiments on multiple benchmarks. Our model achieves competitive performance on the sequence-based action recognition datasets compared to existing clip-based methods. Our model also achieves a significant performance gain on the frame-based multiple object tracking task compared to the previous practice of frame-based models. To the best of our knowledge, this is the first deep learning architecture that unifies video understanding tasks.

In the future, we will apply S-ViT to more video tasks including single object tracking, video object detection, and long-term video localization. Besides, we will continue to improve S-ViT by upgrading its components, such as the detection head. 


\noindent\textbf{Acknowledgement} This work was partially supported by National Key R\&D Program of China under Grant 2020AAA0105702, National Natural Science Foundation of China (NSFC) under Grants 62225207 and U19B2038, and the University Synergy Innovation Program of Anhui Province under Grants GXXT-2019-025.
\newpage
{\small
\bibliographystyle{ieee_fullname}
\bibliography{egbib}

\begin{thebibliography}{10}\itemsep=-1pt

\bibitem{Arnab_2021_ICCV}
Anurag Arnab, Mostafa Dehghani, Georg Heigold, Chen Sun, Mario Lu\v{c}i\'c, and
  Cordelia Schmid.
\newblock Vivit: A video vision transformer.
\newblock In {\em Proceedings of the IEEE/CVF International Conference on
  Computer Vision (ICCV)}, pages 6836--6846, October 2021.

\bibitem{DBLP:journals/ejivp/BernardinS08}
Keni Bernardin and Rainer Stiefelhagen.
\newblock Evaluating multiple object tracking performance: The {CLEAR} {MOT}
  metrics.
\newblock {\em {EURASIP} J. Image Video Process.}, 2008, 2008.

\bibitem{DBLP:conf/icml/BertasiusWT21}
Gedas Bertasius, Heng Wang, and Lorenzo Torresani.
\newblock Is space-time attention all you need for video understanding?
\newblock In {\em {ICML}}, volume 139 of {\em Proceedings of Machine Learning
  Research}, pages 813--824. {PMLR}, 2021.

\bibitem{DBLP:conf/icip/BewleyGORU16}
Alex Bewley, ZongYuan Ge, Lionel Ott, Fabio~Tozeto Ramos, and Ben Upcroft.
\newblock Simple online and realtime tracking.
\newblock In {\em {ICIP}}, pages 3464--3468. {IEEE}, 2016.

\bibitem{DBLP:journals/corr/abs-2004-10934}
Alexey Bochkovskiy, Chien{-}Yao Wang, and Hong{-}Yuan~Mark Liao.
\newblock Yolov4: Optimal speed and accuracy of object detection.
\newblock {\em CoRR}, abs/2004.10934, 2020.

\bibitem{DBLP:journals/corr/abs-2106-05968}
Adrian Bulat, Juan{-}Manuel Perez{-}Rua, Swathikiran Sudhakaran, Brais
  Mart{\'{\i}}nez, and Georgios Tzimiropoulos.
\newblock Space-time mixing attention for video transformer.
\newblock {\em CoRR}, abs/2106.05968, 2021.

\bibitem{DBLP:conf/cvpr/CaiX0XXTS22}
Jiarui Cai, Mingze Xu, Wei Li, Yuanjun Xiong, Wei Xia, Zhuowen Tu, and Stefano
  Soatto.
\newblock Memot: Multi-object tracking with memory.
\newblock In {\em {CVPR}}, pages 8080--8090. {IEEE}, 2022.

\bibitem{DBLP:conf/eccv/CarionMSUKZ20}
Nicolas Carion, Francisco Massa, Gabriel Synnaeve, Nicolas Usunier, Alexander
  Kirillov, and Sergey Zagoruyko.
\newblock End-to-end object detection with transformers.
\newblock In {\em {ECCV} {(1)}}, volume 12346 of {\em Lecture Notes in Computer
  Science}, pages 213--229. Springer, 2020.

\bibitem{DBLP:conf/cvpr/CarreiraZ17}
Jo{\~{a}}o Carreira and Andrew Zisserman.
\newblock Quo vadis, action recognition? {A} new model and the kinetics
  dataset.
\newblock In {\em {CVPR}}, pages 4724--4733. {IEEE} Computer Society, 2017.

\bibitem{DBLP:conf/nips/ChengSK21}
Bowen Cheng, Alexander~G. Schwing, and Alexander Kirillov.
\newblock Per-pixel classification is not all you need for semantic
  segmentation.
\newblock In {\em NeurIPS}, pages 17864--17875, 2021.

\bibitem{DBLP:conf/eccv/ChengS22}
Ho~Kei Cheng and Alexander~G. Schwing.
\newblock Xmem: Long-term video object segmentation with an atkinson-shiffrin
  memory model.
\newblock In {\em {ECCV} {(28)}}, volume 13688 of {\em Lecture Notes in
  Computer Science}, pages 640--658. Springer, 2022.

\bibitem{DBLP:journals/corr/abs-2104-00194}
Peng Chu, Jiang Wang, Quanzeng You, Haibin Ling, and Zicheng Liu.
\newblock Transmot: Spatial-temporal graph transformer for multiple object
  tracking.
\newblock {\em CoRR}, abs/2104.00194, 2021.

\bibitem{DBLP:conf/nips/CubukZS020}
Ekin~Dogus Cubuk, Barret Zoph, Jon Shlens, and Quoc Le.
\newblock Randaugment: Practical automated data augmentation with a reduced
  search space.
\newblock In {\em NeurIPS}, 2020.

\bibitem{DBLP:conf/iclr/DosovitskiyB0WZ21}
Alexey Dosovitskiy, Lucas Beyer, Alexander Kolesnikov, Dirk Weissenborn,
  Xiaohua Zhai, Thomas Unterthiner, Mostafa Dehghani, Matthias Minderer, Georg
  Heigold, Sylvain Gelly, Jakob Uszkoreit, and Neil Houlsby.
\newblock An image is worth 16x16 words: Transformers for image recognition at
  scale.
\newblock In {\em {ICLR}}. OpenReview.net, 2021.

\bibitem{DBLP:conf/iccv/FabbriBMCGOCLC21}
Matteo Fabbri, Guillem Bras{\'{o}}, Gianluca Maugeri, Orcun Cetintas, Riccardo
  Gasparini, Aljosa Osep, Simone Calderara, Laura Leal{-}Taix{\'{e}}, and Rita
  Cucchiara.
\newblock Motsynth: How can synthetic data help pedestrian detection and
  tracking?
\newblock In {\em {ICCV}}, pages 10829--10839. {IEEE}, 2021.

\bibitem{Fan_2021_ICCV}
Haoqi Fan, Bo Xiong, Karttikeya Mangalam, Yanghao Li, Zhicheng Yan, Jitendra
  Malik, and Christoph Feichtenhofer.
\newblock Multiscale vision transformers.
\newblock In {\em Proceedings of the IEEE/CVF International Conference on
  Computer Vision (ICCV)}, pages 6824--6835, October 2021.

\bibitem{DBLP:conf/cvpr/Feichtenhofer20}
Christoph Feichtenhofer.
\newblock {X3D:} expanding architectures for efficient video recognition.
\newblock In {\em {CVPR}}, pages 200--210. Computer Vision Foundation / {IEEE},
  2020.

\bibitem{DBLP:conf/iccv/Feichtenhofer0M19}
Christoph Feichtenhofer, Haoqi Fan, Jitendra Malik, and Kaiming He.
\newblock Slowfast networks for video recognition.
\newblock In {\em {ICCV}}, pages 6201--6210. {IEEE}, 2019.

\bibitem{DBLP:journals/corr/abs-2107-08430}
Zheng Ge, Songtao Liu, Feng Wang, Zeming Li, and Jian Sun.
\newblock {YOLOX:} exceeding {YOLO} series in 2021.
\newblock {\em CoRR}, abs/2107.08430, 2021.

\bibitem{DBLP:conf/iccv/GoyalKMMWKHFYMH17}
Raghav Goyal, Samira~Ebrahimi Kahou, Vincent Michalski, Joanna Materzynska,
  Susanne Westphal, Heuna Kim, Valentin Haenel, Ingo Fr{\"{u}}nd, Peter
  Yianilos, Moritz Mueller{-}Freitag, Florian Hoppe, Christian Thurau, Ingo
  Bax, and Roland Memisevic.
\newblock The "something something" video database for learning and evaluating
  visual common sense.
\newblock In {\em {ICCV}}, pages 5843--5851. {IEEE} Computer Society, 2017.

\bibitem{DBLP:conf/cvpr/HeZRS16}
Kaiming He, Xiangyu Zhang, Shaoqing Ren, and Jian Sun.
\newblock Deep residual learning for image recognition.
\newblock In {\em {CVPR}}, pages 770--778. {IEEE} Computer Society, 2016.

\bibitem{DBLP:conf/cvpr/HofferBHGHS20}
Elad Hoffer, Tal Ben{-}Nun, Itay Hubara, Niv Giladi, Torsten Hoefler, and
  Daniel Soudry.
\newblock Augment your batch: Improving generalization through instance
  repetition.
\newblock In {\em {CVPR}}, pages 8126--8135. Computer Vision Foundation /
  {IEEE}, 2020.

\bibitem{DBLP:journals/corr/KayCSZHVVGBNSZ17}
Will Kay, Jo{\~{a}}o Carreira, Karen Simonyan, Brian Zhang, Chloe Hillier,
  Sudheendra Vijayanarasimhan, Fabio Viola, Tim Green, Trevor Back, Paul
  Natsev, Mustafa Suleyman, and Andrew Zisserman.
\newblock The kinetics human action video dataset.
\newblock {\em CoRR}, abs/1705.06950, 2017.

\bibitem{DBLP:conf/cvpr/KondratyukYLZTB21}
Dan Kondratyuk, Liangzhe Yuan, Yandong Li, Li Zhang, Mingxing Tan, Matthew
  Brown, and Boqing Gong.
\newblock Movinets: Mobile video networks for efficient video recognition.
\newblock In {\em {CVPR}}, pages 16020--16030. Computer Vision Foundation /
  {IEEE}, 2021.

\bibitem{DBLP:conf/eccv/KwonKKC20}
Heeseung Kwon, Manjin Kim, Suha Kwak, and Minsu Cho.
\newblock Motionsqueeze: Neural motion feature learning for video
  understanding.
\newblock In {\em {ECCV} {(16)}}, volume 12361 of {\em Lecture Notes in
  Computer Science}, pages 345--362. Springer, 2020.

\bibitem{Kwon_2021_ICCV}
Heeseung Kwon, Manjin Kim, Suha Kwak, and Minsu Cho.
\newblock Learning self-similarity in space and time as generalized motion for
  video action recognition.
\newblock In {\em ICCV}, 2021.

\bibitem{DBLP:conf/iclr/Li00S00022}
Kunchang Li, Yali Wang, Peng Gao, Guanglu Song, Yu Liu, Hongsheng Li, and Yu
  Qiao.
\newblock Uniformer: Unified transformer for efficient spatial-temporal
  representation learning.
\newblock In {\em {ICLR}}. OpenReview.net, 2022.

\bibitem{DBLP:conf/eccv/LiMGH22}
Yanghao Li, Hanzi Mao, Ross~B. Girshick, and Kaiming He.
\newblock Exploring plain vision transformer backbones for object detection.
\newblock In {\em {ECCV} {(9)}}, volume 13669 of {\em Lecture Notes in Computer
  Science}, pages 280--296. Springer, 2022.

\bibitem{DBLP:journals/corr/abs-2112-01526}
Yanghao Li, Chao{-}Yuan Wu, Haoqi Fan, Karttikeya Mangalam, Bo Xiong, Jitendra
  Malik, and Christoph Feichtenhofer.
\newblock Improved multiscale vision transformers for classification and
  detection.
\newblock {\em CoRR}, abs/2112.01526, 2021.

\bibitem{DBLP:conf/iccv/LinGH19}
Ji Lin, Chuang Gan, and Song Han.
\newblock {TSM:} temporal shift module for efficient video understanding.
\newblock In {\em {ICCV}}, pages 7082--7092. {IEEE}, 2019.

\bibitem{DBLP:journals/corr/abs-2112-00995}
Liting Lin, Heng Fan, Yong Xu, and Haibin Ling.
\newblock Swintrack: {A} simple and strong baseline for transformer tracking.
\newblock {\em CoRR}, abs/2112.00995, 2021.

\bibitem{DBLP:conf/eccv/LinMBHPRDZ14}
Tsung{-}Yi Lin, Michael Maire, Serge~J. Belongie, James Hays, Pietro Perona,
  Deva Ramanan, Piotr Doll{\'{a}}r, and C.~Lawrence Zitnick.
\newblock Microsoft {COCO:} common objects in context.
\newblock In {\em {ECCV} {(5)}}, volume 8693 of {\em Lecture Notes in Computer
  Science}, pages 740--755. Springer, 2014.

\bibitem{DBLP:journals/corr/abs-2208-03550}
Ziyi Lin, Shijie Geng, Renrui Zhang, Peng Gao, Gerard de Melo, Xiaogang Wang,
  Jifeng Dai, Yu Qiao, and Hongsheng Li.
\newblock Frozen {CLIP} models are efficient video learners.
\newblock {\em CoRR}, abs/2208.03550, 2022.

\bibitem{DBLP:conf/cvpr/LiuQQSJ18}
Shu Liu, Lu Qi, Haifang Qin, Jianping Shi, and Jiaya Jia.
\newblock Path aggregation network for instance segmentation.
\newblock In {\em {CVPR}}, pages 8759--8768. Computer Vision Foundation /
  {IEEE} Computer Society, 2018.

\bibitem{Liu_2021_ICCV}
Ze Liu, Yutong Lin, Yue Cao, Han Hu, Yixuan Wei, Zheng Zhang, Stephen Lin, and
  Baining Guo.
\newblock Swin transformer: Hierarchical vision transformer using shifted
  windows.
\newblock In {\em Proceedings of the IEEE/CVF International Conference on
  Computer Vision (ICCV)}, pages 10012--10022, October 2021.

\bibitem{DBLP:journals/corr/abs-2106-13230}
Ze Liu, Jia Ning, Yue Cao, Yixuan Wei, Zheng Zhang, Stephen Lin, and Han Hu.
\newblock Video swin transformer.
\newblock {\em CoRR}, abs/2106.13230, 2021.

\bibitem{DBLP:conf/iclr/LoshchilovH19}
Ilya Loshchilov and Frank Hutter.
\newblock Decoupled weight decay regularization.
\newblock In {\em {ICLR} (Poster)}. OpenReview.net, 2019.

\bibitem{DBLP:journals/ijcv/LuitenODTGLL21}
Jonathon Luiten, Aljosa Osep, Patrick Dendorfer, Philip H.~S. Torr, Andreas
  Geiger, Laura Leal{-}Taix{\'{e}}, and Bastian Leibe.
\newblock {HOTA:} {A} higher order metric for evaluating multi-object tracking.
\newblock {\em Int. J. Comput. Vis.}, 129(2):548--578, 2021.

\bibitem{DBLP:conf/cvpr/MeinhardtKLF22}
Tim Meinhardt, Alexander Kirillov, Laura Leal{-}Taix{\'{e}}, and Christoph
  Feichtenhofer.
\newblock Trackformer: Multi-object tracking with transformers.
\newblock In {\em {CVPR}}, pages 8834--8844. {IEEE}, 2022.

\bibitem{MOT16}
A. Milan, L. Leal-Taix\'{e}, I. Reid, S. Roth, and K. Schindler.
\newblock {MOT}16: {A} benchmark for multi-object tracking.
\newblock {\em arXiv:1603.00831 [cs]}, Mar. 2016.
\newblock arXiv: 1603.00831.

\bibitem{DBLP:journals/corr/abs-2208-02816}
Bolin Ni, Houwen Peng, Minghao Chen, Songyang Zhang, Gaofeng Meng, Jianlong Fu,
  Shiming Xiang, and Haibin Ling.
\newblock Expanding language-image pretrained models for general video
  recognition.
\newblock {\em CoRR}, abs/2208.02816, 2022.

\bibitem{DBLP:journals/corr/abs-2206-13559}
Junting Pan, Ziyi Lin, Xiatian Zhu, Jing Shao, and Hongsheng Li.
\newblock St-adapter: Parameter-efficient image-to-video transfer learning for
  action recognition.
\newblock {\em CoRR}, abs/2206.13559, 2022.

\bibitem{DBLP:conf/cvpr/PangQLCLDY21}
Jiangmiao Pang, Linlu Qiu, Xia Li, Haofeng Chen, Qi Li, Trevor Darrell, and
  Fisher Yu.
\newblock Quasi-dense similarity learning for multiple object tracking.
\newblock In {\em {CVPR}}, pages 164--173. Computer Vision Foundation / {IEEE},
  2021.

\bibitem{DBLP:conf/nips/PatrickCAMMFVH21}
Mandela Patrick, Dylan Campbell, Yuki~M. Asano, Ishan Misra, Florian Metze,
  Christoph Feichtenhofer, Andrea Vedaldi, and Jo{\~{a}}o~F. Henriques.
\newblock Keeping your eye on the ball: Trajectory attention in video
  transformers.
\newblock In {\em NeurIPS}, pages 12493--12506, 2021.

\bibitem{DBLP:conf/iccv/QiuYM17}
Zhaofan Qiu, Ting Yao, and Tao Mei.
\newblock Learning spatio-temporal representation with pseudo-3d residual
  networks.
\newblock In {\em {ICCV}}, pages 5534--5542. {IEEE} Computer Society, 2017.

\bibitem{DBLP:conf/icml/RadfordKHRGASAM21}
Alec Radford, Jong~Wook Kim, Chris Hallacy, Aditya Ramesh, Gabriel Goh,
  Sandhini Agarwal, Girish Sastry, Amanda Askell, Pamela Mishkin, Jack Clark,
  Gretchen Krueger, and Ilya Sutskever.
\newblock Learning transferable visual models from natural language
  supervision.
\newblock In {\em {ICML}}, volume 139 of {\em Proceedings of Machine Learning
  Research}, pages 8748--8763. {PMLR}, 2021.

\bibitem{DBLP:conf/nips/RyooPADA21}
Michael~S. Ryoo, A.~J. Piergiovanni, Anurag Arnab, Mostafa Dehghani, and Anelia
  Angelova.
\newblock Tokenlearner: Adaptive space-time tokenization for videos.
\newblock In {\em NeurIPS}, pages 12786--12797, 2021.

\bibitem{DBLP:journals/corr/abs-1805-00123}
Shuai Shao, Zijian Zhao, Boxun Li, Tete Xiao, Gang Yu, Xiangyu Zhang, and Jian
  Sun.
\newblock Crowdhuman: {A} benchmark for detecting human in a crowd.
\newblock {\em CoRR}, abs/1805.00123, 2018.

\bibitem{DBLP:journals/corr/abs-2012-15460}
Peize Sun, Yi Jiang, Rufeng Zhang, Enze Xie, Jinkun Cao, Xinting Hu, Tao Kong,
  Zehuan Yuan, Changhu Wang, and Ping Luo.
\newblock Transtrack: Multiple-object tracking with transformer.
\newblock {\em CoRR}, abs/2012.15460, 2020.

\bibitem{DBLP:conf/iccv/TranBFTP15}
Du Tran, Lubomir~D. Bourdev, Rob Fergus, Lorenzo Torresani, and Manohar Paluri.
\newblock Learning spatiotemporal features with 3d convolutional networks.
\newblock In {\em {ICCV}}, pages 4489--4497. {IEEE} Computer Society, 2015.

\bibitem{DBLP:conf/iccv/TranWFT19}
Du Tran, Heng Wang, Matt Feiszli, and Lorenzo Torresani.
\newblock Video classification with channel-separated convolutional networks.
\newblock In {\em {ICCV}}, pages 5551--5560. {IEEE}, 2019.

\bibitem{DBLP:conf/cvpr/TranWTRLP18}
Du Tran, Heng Wang, Lorenzo Torresani, Jamie Ray, Yann LeCun, and Manohar
  Paluri.
\newblock A closer look at spatiotemporal convolutions for action recognition.
\newblock In {\em {CVPR}}, pages 6450--6459. Computer Vision Foundation /
  {IEEE} Computer Society, 2018.

\bibitem{DBLP:conf/ssw/OordDZSVGKSK16}
A{\"{a}}ron van~den Oord, Sander Dieleman, Heiga Zen, Karen Simonyan, Oriol
  Vinyals, Alex Graves, Nal Kalchbrenner, Andrew~W. Senior, and Koray
  Kavukcuoglu.
\newblock Wavenet: {A} generative model for raw audio.
\newblock In {\em {SSW}}, page 125. {ISCA}, 2016.

\bibitem{DBLP:conf/nips/VaswaniSPUJGKP17}
Ashish Vaswani, Noam Shazeer, Niki Parmar, Jakob Uszkoreit, Llion Jones,
  Aidan~N. Gomez, Lukasz Kaiser, and Illia Polosukhin.
\newblock Attention is all you need.
\newblock In {\em {NIPS}}, pages 5998--6008, 2017.

\bibitem{DBLP:conf/cvpr/WangLWCHY20}
Chien{-}Yao Wang, Hong{-}Yuan~Mark Liao, Yueh{-}Hua Wu, Ping{-}Yang Chen,
  Jun{-}Wei Hsieh, and I{-}Hau Yeh.
\newblock Cspnet: {A} new backbone that can enhance learning capability of
  {CNN}.
\newblock In {\em {CVPR} Workshops}, pages 1571--1580. Computer Vision
  Foundation / {IEEE}, 2020.

\bibitem{DBLP:journals/corr/abs-2209-07526}
Junke Wang, Dongdong Chen, Zuxuan Wu, Chong Luo, Luowei Zhou, Yucheng Zhao,
  Yujia Xie, Ce Liu, Yu{-}Gang Jiang, and Lu Yuan.
\newblock Omnivl: One foundation model for image-language and video-language
  tasks.
\newblock {\em CoRR}, abs/2209.07526, 2022.

\bibitem{DBLP:conf/cvpr/0002TJW21}
Limin Wang, Zhan Tong, Bin Ji, and Gangshan Wu.
\newblock {TDN:} temporal difference networks for efficient action recognition.
\newblock In {\em {CVPR}}, pages 1895--1904. Computer Vision Foundation /
  {IEEE}, 2021.

\bibitem{DBLP:journals/corr/abs-2109-08472}
Mengmeng Wang, Jiazheng Xing, and Yong Liu.
\newblock Actionclip: {A} new paradigm for video action recognition.
\newblock {\em CoRR}, abs/2109.08472, 2021.

\bibitem{DBLP:conf/cvpr/WangZPX21}
Qiang Wang, Yun Zheng, Pan Pan, and Yinghui Xu.
\newblock Multiple object tracking with correlation learning.
\newblock In {\em {CVPR}}, pages 3876--3886. Computer Vision Foundation /
  {IEEE}, 2021.

\bibitem{DBLP:conf/cvpr/WuF0HKG19}
Chao{-}Yuan Wu, Christoph Feichtenhofer, Haoqi Fan, Kaiming He, Philipp
  Kr{\"{a}}henb{\"{u}}hl, and Ross~B. Girshick.
\newblock Long-term feature banks for detailed video understanding.
\newblock In {\em {CVPR}}, pages 284--293. Computer Vision Foundation / {IEEE},
  2019.

\bibitem{DBLP:conf/cvpr/WuLM0XMF22}
Chao{-}Yuan Wu, Yanghao Li, Karttikeya Mangalam, Haoqi Fan, Bo Xiong, Jitendra
  Malik, and Christoph Feichtenhofer.
\newblock Memvit: Memory-augmented multiscale vision transformer for efficient
  long-term video recognition.
\newblock In {\em {CVPR}}, pages 13577--13587. {IEEE}, 2022.

\bibitem{DBLP:conf/cvpr/WuCS00Y21}
Jialian Wu, Jiale Cao, Liangchen Song, Yu Wang, Ming Yang, and Junsong Yuan.
\newblock Track to detect and segment: An online multi-object tracker.
\newblock In {\em {CVPR}}, pages 12352--12361. Computer Vision Foundation /
  {IEEE}, 2021.

\bibitem{DBLP:journals/corr/abs-2207-01297}
Wenhao Wu, Zhun Sun, and Wanli Ouyang.
\newblock Transferring textual knowledge for visual recognition.
\newblock {\em CoRR}, abs/2207.01297, 2022.

\bibitem{DBLP:conf/eccv/XieSHTM18}
Saining Xie, Chen Sun, Jonathan Huang, Zhuowen Tu, and Kevin Murphy.
\newblock Rethinking spatiotemporal feature learning: Speed-accuracy trade-offs
  in video classification.
\newblock In {\em {ECCV} {(15)}}, volume 11219 of {\em Lecture Notes in
  Computer Science}, pages 318--335. Springer, 2018.

\bibitem{DBLP:conf/iccv/Xu0ZH19}
Jiarui Xu, Yue Cao, Zheng Zhang, and Han Hu.
\newblock Spatial-temporal relation networks for multi-object tracking.
\newblock In {\em {ICCV}}, pages 3987--3997. {IEEE}, 2019.

\bibitem{DBLP:journals/corr/abs-2103-15145}
Yihong Xu, Yutong Ban, Guillaume Delorme, Chuang Gan, Daniela Rus, and Xavier
  Alameda{-}Pineda.
\newblock Transcenter: Transformers with dense queries for multiple-object
  tracking.
\newblock {\em CoRR}, abs/2103.15145, 2021.

\bibitem{DBLP:conf/eccv/YanJSWYLL22}
Bin Yan, Yi Jiang, Peize Sun, Dong Wang, Zehuan Yuan, Ping Luo, and Huchuan Lu.
\newblock Towards grand unification of object tracking.
\newblock In {\em {ECCV} {(21)}}, volume 13681 of {\em Lecture Notes in
  Computer Science}, pages 733--751. Springer, 2022.

\bibitem{DBLP:journals/corr/abs-2201-04288}
Shen Yan, Xuehan Xiong, Anurag Arnab, Zhichao Lu, Mi Zhang, Chen Sun, and
  Cordelia Schmid.
\newblock Multiview transformers for video recognition.
\newblock {\em CoRR}, abs/2201.04288, 2022.

\bibitem{DBLP:conf/iccv/YunHCOYC19}
Sangdoo Yun, Dongyoon Han, Sanghyuk Chun, Seong~Joon Oh, Youngjoon Yoo, and
  Junsuk Choe.
\newblock Cutmix: Regularization strategy to train strong classifiers with
  localizable features.
\newblock In {\em {ICCV}}, pages 6022--6031. {IEEE}, 2019.

\bibitem{DBLP:conf/eccv/ZengDZWZW22}
Fangao Zeng, Bin Dong, Yuang Zhang, Tiancai Wang, Xiangyu Zhang, and Yichen
  Wei.
\newblock {MOTR:} end-to-end multiple-object tracking with transformer.
\newblock In {\em {ECCV} {(27)}}, volume 13687 of {\em Lecture Notes in
  Computer Science}, pages 659--675. Springer, 2022.

\bibitem{DBLP:conf/iclr/ZhangCDL18}
Hongyi Zhang, Moustapha Ciss{\'{e}}, Yann~N. Dauphin, and David Lopez{-}Paz.
\newblock mixup: Beyond empirical risk minimization.
\newblock In {\em {ICLR} (Poster)}. OpenReview.net, 2018.

\bibitem{DBLP:conf/eccv/ZhangSJYWYLLW22}
Yifu Zhang, Peize Sun, Yi Jiang, Dongdong Yu, Fucheng Weng, Zehuan Yuan, Ping
  Luo, Wenyu Liu, and Xinggang Wang.
\newblock Bytetrack: Multi-object tracking by associating every detection box.
\newblock In {\em {ECCV} {(22)}}, volume 13682 of {\em Lecture Notes in
  Computer Science}, pages 1--21. Springer, 2022.

\bibitem{DBLP:journals/ijcv/ZhangWWZL21}
Yifu Zhang, Chunyu Wang, Xinggang Wang, Wenjun Zeng, and Wenyu Liu.
\newblock Fairmot: On the fairness of detection and re-identification in
  multiple object tracking.
\newblock {\em Int. J. Comput. Vis.}, 129(11):3069--3087, 2021.

\bibitem{zhao2023td}
Yucheng Zhao, Chong Luo, Chuanxin Tang, Dongdong Chen, Noel~C Codella, Lu Yuan,
  and Zheng-Jun Zha.
\newblock T2d: Spatiotemporal feature learning based on triple 2d
  decomposition, 2023.

\bibitem{DBLP:conf/aaai/Zhong0KL020}
Zhun Zhong, Liang Zheng, Guoliang Kang, Shaozi Li, and Yi Yang.
\newblock Random erasing data augmentation.
\newblock In {\em {AAAI}}, pages 13001--13008. {AAAI} Press, 2020.

\bibitem{DBLP:conf/eccv/ZhouKK20}
Xingyi Zhou, Vladlen Koltun, and Philipp Kr{\"{a}}henb{\"{u}}hl.
\newblock Tracking objects as points.
\newblock In {\em {ECCV} {(4)}}, volume 12349 of {\em Lecture Notes in Computer
  Science}, pages 474--490. Springer, 2020.

\bibitem{DBLP:conf/cvpr/ZhouYKK22}
Xingyi Zhou, Tianwei Yin, Vladlen Koltun, and Philipp Kr{\"{a}}henb{\"{u}}hl.
\newblock Global tracking transformers.
\newblock In {\em {CVPR}}, pages 8761--8770. {IEEE}, 2022.

\bibitem{DBLP:conf/eccv/ZolfaghariSB18}
Mohammadreza Zolfaghari, Kamaljeet Singh, and Thomas Brox.
\newblock {ECO:} efficient convolutional network for online video
  understanding.
\newblock In {\em {ECCV} {(2)}}, volume 11206 of {\em Lecture Notes in Computer
  Science}, pages 713--730. Springer, 2018.

\end{thebibliography}
}

\end{document}


\title{Streaming Video Model \\Supplementary Materials}

\author{
Yucheng Zhao$^{1}$\footnotemark[1], Chong Luo$^{2}$, Chuanxin Tang$^{2}$, Dongdong Chen$^{3}$, Noel Codella$^{3}$, Zheng-Jun Zha$^{1}$\footnotemark[2] \\
$^{1}$University of Science and Technology of China \quad $^{2}$Microsoft Research Asia \quad
$^{3}$Microsoft Cloud + AI \\
{\tt\small \{lnc@mail., zhazj\}@ustc.edu.cn \quad \{cluo,chutan,dochen,ncodella\}@microsoft.com}
}

\maketitle

\renewcommand{\thefootnote}{\fnsymbol{footnote}}
\footnotetext[1]{This work was done during the internship of Yucheng at MSRA.}
\footnotetext[2]{Corresponding author.}
\renewcommand{\thefootnote}{\arabic{footnote}}

\appendix

\newcommand{\beginsupplement}{%
        \setcounter{table}{0}
        \renewcommand{\thetable}{S\arabic{table}}%
        \setcounter{figure}{0}
        \renewcommand{\thefigure}{S\arabic{figure}}%
     }
\beginsupplement

\section{Hyperparameter Details}

We present the training hyper-parameters used in different datasets in Tab.\ref{tab:s1} and Tab.\ref{tab:s2}. The hyper-parameters for action recognition are adapted from T2D \cite{zhao2023td}. And the hyper-parameters for multiple object tracking are adapted from ByteTrack \cite{DBLP:conf/eccv/ZhangSJYWYLLW22}. The learning rates shown in Tables are for CLIP pre-trained parameters. For randomly initialized parameters, we use a 100$\times$ learning rate for K400 and a 10$\times$ learning rate for SSv2 and MOT17. In MOT training, we apply sequence-level data augmentation, which means the random parameters are kept the same for frames from the same training sequence. Each training sample is sampled from a video sequence with a frame length of 2 and a random frame interval sampled from $\{1, 2, 3 ,4\}$.

\begin{table}[tb]
  \centering
  \caption{Hyper-parameters used in the action recognition.}
    \begin{tabu}{@{}lcc@{}}
    \toprule
    Dataset & K400 & SSv2 \\
    \hline
    Batch size & 256 & 64\\
    Epochs  & \multicolumn{2}{c}{30}\\
    Warmup epochs  & \multicolumn{2}{c}{5}\\
    Learning rate  & 1e-5 & 5e-5 \\
    Learning rate schedule  & \multicolumn{2}{c}{cosine}\\
    Optimizer & \multicolumn{2}{c}{AdamW}\\
    Weight decay  & 1e-3 & 5e-2\\
    \hline 
    RandomFlip & 0.5 & -\\
    ColorJitter     &0.8 & -\\
    GrayScale &     0.2 & -\\
    RandomAugment   &-   & \checkmark \\
    Random erasing  & -  & 0.25 \\
    Repeated augmentation & - & 2 \\
    Label smoothing &\multicolumn{2}{c}{0.1}\\
    Mixup &\multicolumn{2}{c}{0.8}\\
    CutMix &\multicolumn{2}{c}{1.0}\\
    \bottomrule
  \end{tabu}
  \label{tab:s1}
\end{table}

\begin{table}[tb]
  \centering
  \caption{Hyper-parameters used in the multiple object tracking.}
    \begin{tabu}{@{}lc@{}}
    \toprule
    Dataset & MOT17 \\
    \hline
    Batch size & 16\\
    Epochs  & 10\\
    Warmup epochs  & 1\\
    Learning rate  & 2.5e-5 \\
    Learning rate schedule  & cosine\\
    Optimizer & AdamW\\
    Weight decay  & 0.05\\
    \hline 
    RandomFlip & 0.5 \\
    Mosaic & \checkmark\\
    RandomAffine & \checkmark \\
    RandomHSV & \checkmark \\
    Mixup & \checkmark\\
    \bottomrule
  \end{tabu}
  \label{tab:s2}
\end{table}

\section{Additional Experiments}

we perform a number of ablation experiments on the MOT17 half-validation set to substantiate the rationale for certain design decisions

In order to investigate the necessity of a ResNet block in S-ViT for cross-window spatial information propagation, we first conduct an examination. The results in Tab. \ref{tab:s3} demonstrate that when the ResNet blocks in S-ViT are removed, a significant reduction in tracking performance is observed across all scores, providing evidence for the effectiveness of ResNet blocks in this context.

First, we investigate the necessity of the ResNet block in S-ViT for cross-window spatial information propagation. As shown in Tab. \ref{tab:s3}, when we remove the ResNet blocks in S-ViT, a significant performance loss across all tracking scores is observed, demonstrating the effectiveness of using ResNet blocks. 

\begin{table}[h]
    \centering
    \caption{Analysis of the S-ViT architecture. Introducing the ResNet block is crucial to acquiring good performance on MOT. The results are from the MOT17 half-validation set.}
    \begin{tabular}{@{}l|c|c|c@{}}
        \toprule
        Method  & MOTA $\uparrow$& IDF1 $\uparrow$ & HOTA $\uparrow$ \\
        \hline
        w/ ResNet block  & 79.6 & 80.9 & 68.3 \\
        w/o ResNet block & 78.6 & 77.9 & 66.4\\
        \bottomrule
    \end{tabular}
  \label{tab:s3}

\end{table}

Second, we evaluate S-ViT models with various pre-training strategies. Our default model uses CLIP pre-training. Additionally, we test ImageNet 21K pre-training and ImageNet 1K pre-training, as well as training from scratch. The comparison results are shown in Tab.\ref{tab:s4}. When using no pre-training or the weak ImageNet 1K pre-training, our model does not achieve high tracking scores. While a strong pre-trained model like CLIP or ImageNet 21K is utilized, our model could achieve top performance. Another possible pre-training strategy is to use a detection-trained model, which is a more commonly adopted approach by other MOT methods. However, as there is no available out-of-the-box detection checkpoint for our model, we plan to explore this direction in future work.

\begin{table}[h]

    \centering
    \caption{Analysis of the pre-trained model. We use the CLIP pre-trained model in the default setting, which is proven to perform best. The results are from the MOT17 half-validation set.}
    \begin{tabular}{@{}l|c|c|c@{}}
        \toprule
        Pre-trained Model  & MOTA $\uparrow$& IDF1 $\uparrow$ & HOTA $\uparrow$ \\
        \hline
        CLIP (Default) & 79.6 & 80.9 & 68.3 \\
        ImageNet 21K & 79.4 & 79.7 & 67.7 \\
        ImageNet 1K  & 76.1 & 75.3 & 63.7\\
        No pre-train & 76.1 & 76.9 & 65.0 \\
        \bottomrule
    \end{tabular}
      \label{tab:s4}

\end{table}

Third, we investigate an important hyper-parameter, namely the detection score threshold. As the association method we used in S-ViT does not require any training procedure, the selection of an appropriate hyper-parameter is essential for achieving high performance. Through experiments (see Tab.\ref{tab:s5}), we observed that a relatively high detection score threshold is necessary, as adopting the same value of 0.6 as ByteTrack results in a significant loss in performance.  This observation can be attributed to the distinct detection frameworks utilized by ByteTrack and S-ViT, which tend to generate varying levels of confidence for detection boxes.

\begin{table}[h]
    \centering
    \caption{Analysis of the detection score threshold for S-ViT. Due to the usage of different detection frameworks, S-ViT needs to tune a separate detection threshold to work well. The results are from the MOT17 half-validation set.}
    \begin{tabular}{@{}l|c|c|c@{}}
        \toprule
        Detection Threshold  & MOTA $\uparrow$& IDF1 $\uparrow$ & HOTA $\uparrow$ \\
        \hline
        0.75          & 78.9 & 79.6 & 67.6 \\
        0.7 (S-ViT) & 79.6 & 80.9 & 68.3\\
        0.65          & 79.6 & 79.3 & 67.2\\
        0.6 (ByteTrack)           & 78.7 & 78.0 & 66.2\\
        \bottomrule
    \end{tabular}
  \label{tab:s5}
\end{table}

\newpage
{\small
\bibliographystyle{ieee_fullname}
\bibliography{egbib}
}